# Outcome-Based Education: Evaluating Students' Perspectives Using Transformer

Shuvra Smaran Das
ELITE Lab
New York, USA
shuvradas59@gmail.com

Anirban Saha Anik
ELITE Lab
New York, USA
anirbansaha002@gmail.com

Md Kishor Morol
ELITE Lab
New York, USA
kishoremorol@gmail.com

Mohammad Sakib Mahmood
American International
University-Bangladesh
Dhaka, Bangladesh
14-27622-3@student.aiub.edu

*Abstract*—Outcome-Based Education (OBE) emphasizes the development of specific competencies through student-centered learning. In this study, we reviewed the importance of OBE and implemented transformer-based models, particularly DistilBERT, to analyze an NLP dataset that includes student feedback. Our objective is to assess and improve educational outcomes. Our approach is better than other machine learning models because it uses the transformer's deep understanding of language context to classify sentiment better, giving better results across a wider range of matrices. Our work directly contributes to OBE's goal of achieving measurable outcomes by facilitating the identification of patterns in student learning experiences. We have also applied LIME (local interpretable model-agnostic explanations) to make sure that model predictions are clear. This gives us understandable information about how key terms affect sentiment. Our findings indicate that the combination of transformer models and LIME explanations results in a strong and straightforward framework for analyzing student feedback. This aligns more closely with the principles of OBE and ensures the improvement of educational practices through data-driven insights.

## I. INTRODUCTION

Outcome-based education (OBE) is an educational approach that prioritizes the objectives of each component of a learning system. Students receive individualized feedback on the extent to which they have achieved their objectives, as well as regular evaluations of their progress. It can be defined as a pedagogical model that aims to achieve specific learning outcomes at the end of any educational experience. It formed in the 1980s, and this article [1] was the first to advocate for its systemic reform of educational practices to guarantee at least easily measurable outcomes, among all the academics who are currently delving into the subject. The OBE, which originated in the United States and South Africa, had a significant impact on the education reform movements on equity and accountability in the late 1970s and early 1980s [2]. The system has been adopted and is currently in use in Asia, Europe, and Africa [3].

The primary objective of the OBE system is to transition from the input-based education model, which emphasizes the completion of a certain number of instructional hours, to an education model that guarantees that students acquire a specific set of competencies. The student-centered approach is the primary focus of OBE, which emphasizes the abilities that learners possess after the learning process [4]. As long as the learners attain the specified objectives, instructors are permitted to implement various instructional methodologies [5]. In addition, OBE simplifies the process of transparency and accountability by defining and evaluating outcome statements, which in turn predict the educational effectiveness of those involved [6].

OBE is a systematic approach involving curriculum design, instruction, and assessment is necessary. The curriculum should expressly specify the intended outcomes, and the assessments that are administered should quantify these accomplishments. For example, the implementation of the OBE throughout Malaysia's higher education system required a comprehensive overhaul of the curriculum and instructional strategies to guarantee that students' anticipated learning outcomes were met [7]. Additionally, the inclusion of technological innovations in the implementation of OBE has facilitated the monitoring of students' progress toward achieving outcomes [8].

Nevertheless, this article [9] contends that the rigorous emphasis on measurable outcomes in OBE appears to impede educational flexibility and creativity. Additionally, the implementation processes for OBE are resource-intensive. This implies that the development of curricula and the format of assessments in conjunction with predetermined outcomes require a significant amount of time and administrative effort [10]. OBE is essential for the educational requirements of the present day due to its emphasis on the competencies of students and the workforce's fundamental requirements. OBE can enable each educational institution to ensure that its graduates have the ability to meet the real-world requirements of the workforce in terms of knowledge and skills [30].

The quality of education is determined by OBE research, which also identifies learning deficits that need to be addressed. Engaging Natural Language Processing(NLP) allows us to evaluate students' evaluations of the curriculum design and the manner of instruction. Subsequently, institutions will accumulate critical data to assist them in optimizing course design and instructional methodologies. This method ensures the attainment of learning outcomes by guaranteeing continuous improvement in the implementation of OBE. Fundamental components of our daily existence are emotions and opinions. Opinions are sentiments, convictions, or subjective assessments that individuals hold with respect to a particular subject. Opinions in contrast to knowledge-based assertions may exhibit a variety of polarities and lack factual evidence. Various countries' Ministries of Education endeavor to improve the current educational system by integrating innovative curriculum terminologies and conducting research. As a result, it is crucial that students effectively manage and adjust to this situation. Providing the government with both positive and negative feedback can assist in determining whether to maintain or modify the system. We can understand the current situation by analyzing the inquiries and responses of students who are currently enrolled in a newly implemented curriculum. It is imperative to implement data mining methodologies in order to achieve the intended result. A specialized discipline within data mining [31], Educational Data Mining (EDM) is dedicated to the development and application of methodologies and techniques for extracting valuable knowledge or information from unprocessed data. The ultimate objective of EDM is to improve the efficacy of the education system. Educational data mining has been implemented in recent years by educational systems' research.

The implementation of OBE, a student-centric teaching system that emphasizes quantitative measurements of an individual, is intended to guarantee that education is carried out appropriately. The research is conducted by text mining the data collected from the feedback of students and faculty members of a university with the intention of applying a machine learning approach to analyze the sentiment of individuals who are directly related to or have experience with the OBE curriculum, keeping these points in mind. Ultimately, the study will determine which algorithm produces the most accurate output by utilizing a Bag-of-Words (BoW) model to extract features, applying traditional machine learning algorithms such as Naive Bayes, Support Vector Machines, and logistic regression, and evaluating the results based on accuracy, precision, recall, F1 score, and the ability to correctly identify any given sentence. Additionally, we conducted a comparison between these conventional machine learning models and our customizable ML model through the integration of transformers.

NLP simplifies the identification of patterns in students' experiences that may not have been identified through conventional assessments. Also, it enables personalized learning by addressing the unique requirements of each student, thereby enhancing student engagement and satisfaction. In summary, the continuous cycle of development that is established by research into NLP within OBE connects the excessively theoretical educational models to the actual classroom practices.

Our objectives of this research works-
- ➢ Collect and analyze student feedback regarding the OBE system, with a focus on their distinctive

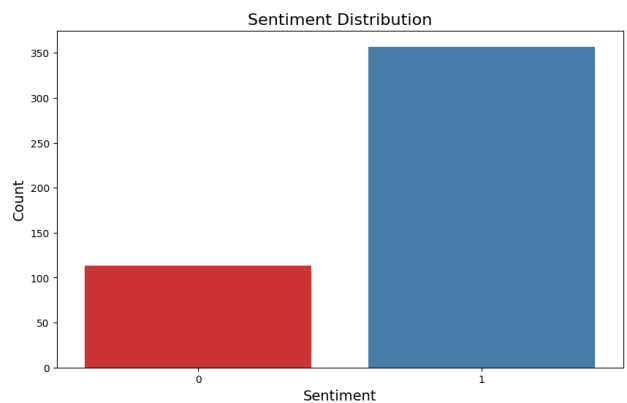

Fig 1. Positive and Negative Sentiment Distribution.

  perspectives and experiences, utilizing advanced NLP techniques.
- ➢ Address the void in current OBE research, which primarily focuses on faculty or instructor feedback, by incorporating the student perspective. This will offer a more thorough comprehension of the OBE system.
- ➢ Establish a framework for analyzing feedback from students and teaching staff, utilizing NLP models like DistilBERT to improve sentiment classification and data analysis and to identify significant trends and patterns in the feedback.
- ➢ Include LIME to guarantee the transparency and interpretability of the machine learning model utilized for analysis. This approach provides explicit explanations for the model's classification and interpretation of the feedback.
- ➢ Enhance the educational experience for students by implementing a more targeted approach to curriculum design and instructional methods, which will contribute to the continuous development of the OBE system by providing data-driven insights based on the collected feedback.

## II. RELATED WORK

Through opinion and text analysis, educational data mining has demonstrated its effectiveness in enhancing curricula. Sentiment analysis of student feedback has been demonstrated to offer valuable insights for the improvement of teaching practices in studies conducted by [32] and [14]. Sentiment analysis, which employs Natural Language Processing (NLP), is a critical approach to monitoring opinions regarding educational subjects, as per [15].

Automating text classification for sentiment detection, a technique that has been extensively employed in educational feedback, was the primary focus of this research paper [16]. The efficacy of machine learning models [12] [13], including deep learning and bundles of words, in the classification and analysis of student feedback is illustrated by studies such as [17] and [18]. In another method, [19] employed NLP to conduct sentiment analysis on student responses in order to enhance the processes of higher education. In the same vein, this paper [20] investigated the use of NLP for feedback analysis both prior to and following the COVID-19 pandemic [11]. This research work [21] also investigates lexicon-based sentiment analysis, which offers an additional approach to analyzing substantial quantities of feedback. However, it necessitates structured input to ensure precision.

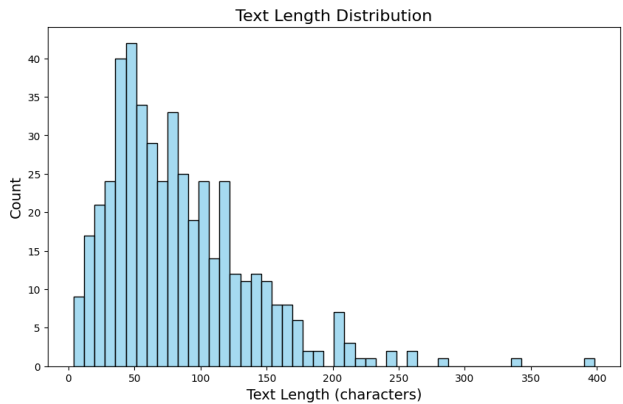

Fig 2. Text Length Distribution.

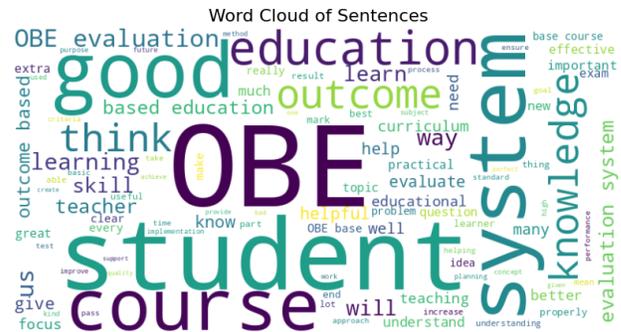

Fig 3. World Clouds of the dataset.

### III. METHODOLOGY

The proposed methodology describes the steps taken in collecting, cleaning, and analyzing data for our sentiment analysis task. First of all, responses for people involved in the OBE course regarding experiences were collected. The obtained data was preprocessed by removing stop words, lemmatization, and tokenization. We compared some traditional machine learning models such as Naive Bias, Random Forest with state-of-the-art transformer-based model, DistilBERT and was fine-tuned especially for sentiment classification, using our dataset. To better understand the model's predictions, we used the LIME technique; this gave insights into the decisions of the model.

#### A. Data Collection & Cleaning

The target population consisted solely of individuals who had completed courses as either students or instructors of this curriculum, as the Outcome-Based Education (OBE) system has not yet been widespread. The questionnaire was open-ended to encourage extensive responses from participants. The data was collected using Google Forms and subsequently converted to Google Sheets. A total of 478 responses from students and one response from a faculty member were collected in Google Forms. These sentences have positive, negative, and suggestive aspects. Dataset: github.com/iamshuvra/OBE

The sentences of the paragraph were segmented to ascertain the sentiment in each one. The labeled sentences were classified as either positive or negative, denoted by '1' or '0'. Responses with absent values are eliminated to maintain the dataset's integrity. Additionally, duplicate rows are detected and eliminated to avert biassed learning.

#### B. Data Pre-processing

Preparing the dataset for model training is the initial phase for our methodology. Initially, we import and examine the dataset, which comprises textual data and associated sentiment labels. We analyzed its structure by presenting summary statistics and verifying the presence of any missing or duplicate values. We initially removed 'Stop Words' from the phrases [29].Stopwords in Technical Language Processing. We used the natural language processing (nltk) Stopwords package to eliminate stop words from the sentences and lemmatized the words to present them in their base form, e.g., teacher, taught, teaching = teach. We identified and eliminated the "stopwords" utilizing the nltk.corpus import stopwords package. This is a compilation of English lexical stop words. Subsequently, we developed a function to tokenize the words by eliminating punctuation and segmenting the phrases by regular expressions.

#### C. Model Training

To perform sentiment analysis, we trained both traditional machine learning models and a transformer-based model, DistilBERT, to compare their performance.

##### 1) Traditional Machine Learning Models

We trained a couple of traditional machine-learning models to compare the results with the transformer.

- Naive Bayes Algorithm: This probabilistic algorithm predicts the output class based on feature independence. We calculate the class probability and choose the one with the highest probability. Various applications, including spam detection, have successfully applied Naive Bayes [22] and [23].
- Logistic Regression: A predictive model based on probability and the sigmoid function. People widely use it for binary classification tasks. Studies like those by [24] demonstrate its performance in text classification.
- Random Forest: is an ensemble technique that grows multiple decision trees and makes predictions based on the majority vote of the trees. It is effective in handling overfitting and high-dimensional data. Predictive analytics has used it for diseases like diabetes and heart disease [25].
- K-Nearest Neighbor (KNN): A non-parametric algorithm that stores the dataset and classifies data based on the majority vote of its neighbors. KNN is simple yet effective for tasks like disease prediction [26].
- Support Vector Machine (SVM): A supervised learning method used for classification and regression,

particularly effective in high-dimensional spaces. Studies have demonstrated SVM's success in tasks such as the classification of chronic diseases [27].

*2) Transformer-Based Model (DistilBERT)*

These models were trained using the preprocessed textual data, and their performance was evaluated using accuracy and other relevant metrics.

We utilized DistilBERT, a compressed variant of the BERT model, for sentiment analysis. DistilBERT is 60% more efficient and has a reduced size compared to BERT. This makes it appropriate for refining our sentiment analysis work. We load the DistilBERT model, pre-trained on an extensive corpus of textual data, using the Transformers library. We further refine the pre-trained model using our specific dataset.

We tokenize the textual data and transform it into a format that is compatible with DistilBERT. We divided the data into training and testing sets, implementing a stratified split to preserve the distribution of sentiment labels. We add a fully connected layer on top of DistilBERT to classify the output into sentiment labels.

The text data is tokenized using the DistilBERT tokenizer and split into training and testing sets. A custom dataset class is defined to handle text tokenization and input preparation. DataLoader is used to efficiently load the data in batches during training.

The model is trained using an AdamW optimizer, which incorporates weight decay to reduce overfitting.

The training method entails forward propagation to calculate the loss and backward propagation to adjust the model weights. We perform the training and evaluation with a batch size of 16. We assess the model's performance by tracking its accuracy and loss during training and validation, which we determine through hyperparameter tuning. As we don't have that large a sample dataset, we have trained our model for 15 epochs.

Furthermore, we assess the model's effectiveness on the test set using accuracy and loss metrics. We used graphs to depict the training and testing loss and accuracy over epochs, demonstrating the model's learning curve.

*3) Explainability with LIME*

To better understand the behavior of our DistilBERT-based sentiment classifier, we used LIME to explain its predictions on both positive and negative sentiment samples from our test dataset. LIME approximates the behavior of a black-box model using a locally interpretable model, offering insights into the model's decision-making process [28]
We initialize the LimeTextExplainer from the LIME library. This explainer produces local explanations by altering the input text and analyzing the model's predictions on these modified samples. For each test instance, we produce LIME explanations that pinpoint the terms most influential to the model's prediction. These elucidations are illustrated to facilitate a comprehensive understanding of the model's behavior.

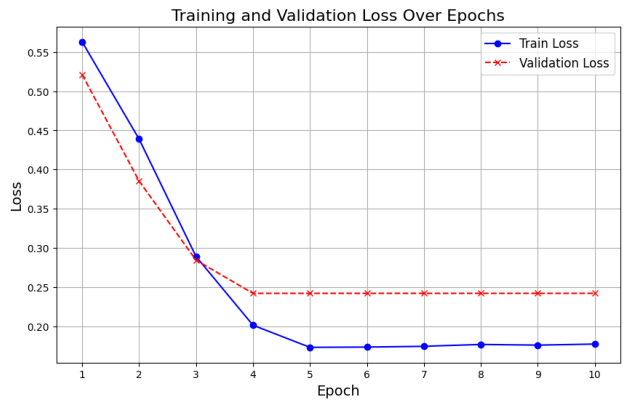

Fig 4. Train and Validation Loss over the number of epochs.

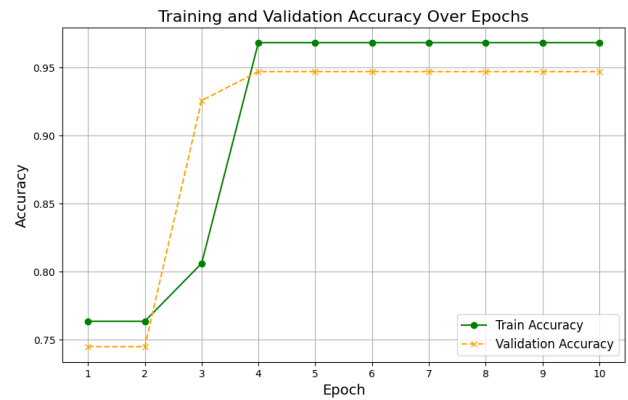

Fig 5. Training and Validation Accuracy Over Epoch.

IV. RESULT AND DISCUSSION

*A. Model Performance*

The performance of the DistilBERT model is assessed based on accuracy and loss metrics on both training and testing datasets. The results demonstrate the model's capability to effectively learn from the data and generalize to unseen examples.

The training and testing loss is plotted across epochs, illustrating how the model converges during training. A steady decrease in loss indicates effective learning, while a stabilization of the loss suggests convergence.

The accuracy of the training and testing datasets is plotted over epochs. High accuracy indicates the model's proficiency in correctly predicting sentiment labels.

*B. Explainability with LIME*

LIME provides local explanations for the model's predictions, offering insights into the features that influence the model's decisions. By approximating the black-box model with an interpretable model locally around the prediction, LIME helps in understanding the importance of individual words in the input text. The LIME explanations for both positive and negative samples indicate that the model significantly relies on sentiment-laden words when making its predictions. Positive sentiments are characterized by the prominence of terms such as "helps," "think," and "effective," while negative sentiments are characterized by words such as "problems," "face," and "not helpful."

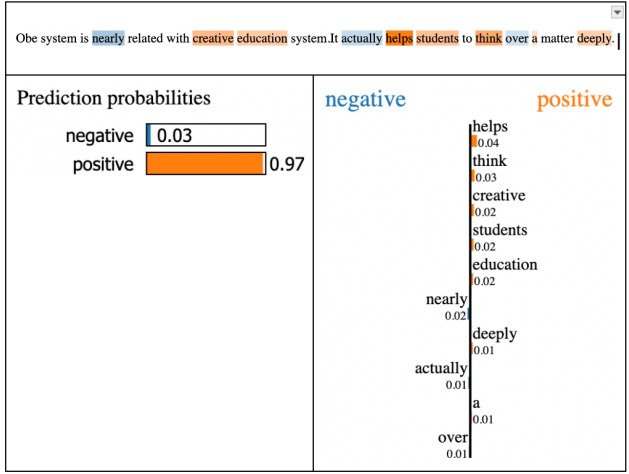

Fig 6. An Example of Positive LIME Explanation where positive words like 'helps', 'think' have more impact.

TABLE I. TABLE OF ACCURACY, PRECISION, RECALL, AND F1 SCORES

| Algorithm | Polarity | Precision | Recall | F1 Score | Accuracy |
|---|---|---|---|---|---|
| Logistic regression | 0 | 0.93 | 0.71 | 0.81 | 0.91 |
|  | 1 | 0.92 | 0.93 | 0.94 |  |
| Naïve Bayes | 0 | 0.92 | 0.57 | 0.71 | 0.88 |
|  | 1 | 0.88 | 0.95 | 0.92 |  |
| Random Forest | 0 | 0.95 | 0.48 | 0.65 | 0.87 |
|  | 1 | 0.87 | 0.94 | 0.90 |  |
| KNN Algorithm | 0 | 0.51 | 0.86 | 0.36 | 0.71 |
|  | 1 | 0.80 | 0.31 | 0.74 |  |
| Support Vector Machine | 0 | 0.64 | 0.76 | 0.70 | 0.83 |
|  | 1 | 0.92 | 0.85 | 0.88 |  |
| **DistilBERT** | **0** | **1.00** | **0.88** | **0.93** | **0.96** |
|  | **1** | **0.96** | **1.0** | **0.98** |  |

Our model using DistilBERT demonstrated (Table I) significantly superior performance in comparison to other machine learning classifiers during our evaluations, which utilized precision, recall, F1-score, and the ROC curve as metrics. DistilBERT achieved an accuracy of 96%, surpassing that of other classifiers, including Naive Bayes (88%) and Logistic Regression (91%).

This level of interpretability(Fig 6, 7) enables us to have confidence that the model is accurately identifying critical features in the text. However, there are also instances in which LIME identifies neutral words that contribute to the sentiment classification, suggesting that there may be area for development in the model's reliance on context.

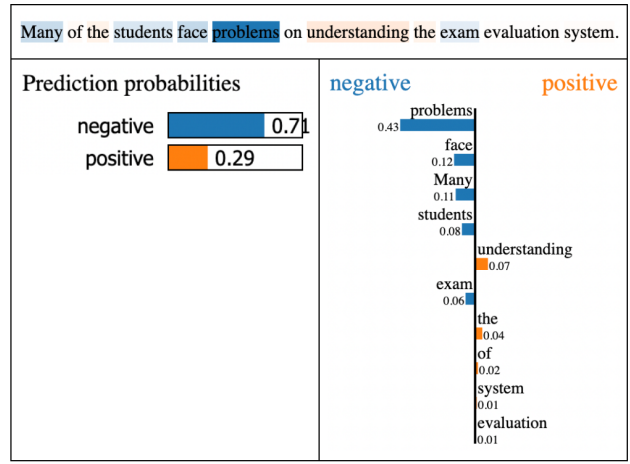

Fig 7. An Example of Negative LIME Explanation where words like 'problems', 'face' have more impact.

For each test instance, LIME highlights the most influential words contributing to the predicted sentiment. These visualizations aid in validating the model's reasoning and identifying any potential biases or errors.

The majority of classifiers, such as DistilBERT, Logistic Regression, and Naive Bayes, accurately predicted basic input sentences such as "**OBE is beneficial for the system**" or "**It is satisfactory**" during the testing process. Nevertheless, DistilBERT was the only traditional classifier that accurately classified the sentiment as negative when presented with more complex and nuanced sentences, such as "**I wish I could say something good but I can't because it's too bad**." This illustrates that transformer-based models, such as DistilBERT, are more adept at managing nuanced language and context, rendering them especially well-suited for sentiment analysis tasks that require subtle word interactions.

## V. CONCLUSION

Outcome-Based Education (OBE) has demonstrated its effectiveness as a reform that prioritizes the development of students' competencies for the 21st-century workforce. OBE promotes flexibility in teaching methods by emphasizing measurable student learning outcomes, while simultaneously demanding careful planning, curriculum design, and assessment strategies for successful implementation. In our investigation, we implemented DistilBERT, a transformer-based model, to evaluate sentiment in relation to the effectiveness of the OBE and analyze student feedback. DistilBERT outperformed conventional machine learning classifiers, including Naive Bayes and logistic regression, in terms of accuracy, precision, recall, and F1-score, achieving 96% accuracy on the validation set. The transformer model was more adept at discerning the emotions of individuals when they utilized complex language, which is more indicative of the way students actually reacted to OBE initiatives, due to its superior comprehension of the context. Additionally, we applied LIME explanations to facilitate comprehension of the model predictions. This made it evident how specific words affected the sentiment classification, which is a very important aspect of educational analysis.

The restricted dataset size was one of the primary obstacles encountered in this investigation. Responses were fewer than anticipated due to the open-ended nature of the questionnaire and a lack of engagement. Additionally, the university conducting this research was relatively new to the concept of OBE, which posed a greater challenge in terms of collecting a wide range of feedback. However, the open-ended approach enabled us to gather a wide variety of responses, providing unique perspectives on students' perceptions of OBE.

The present model is capable of accurately classifying sentiments in declarative sentences; however, it is restricted in its ability to handle imperative and interrogative sentences. Future research should be directed toward the expansion of the dataset and the refinement of the model to more effectively capture these types of sentences, as well as the measurement of positivity or negativity in a more detailed manner. Subsequent development could also investigate the incorporation of sophisticated sentiment analysis techniques to evaluate the intensity and subtleties of sentiments in student feedback, in addition to determining whether they are positive or negative.